\documentclass[10pt,twocolumn,letterpaper]{article}
\usepackage{cvpr}
\usepackage{times}
\usepackage{epsfig}
\usepackage{graphicx}
\usepackage{amsmath}
\usepackage{amssymb}
\usepackage{diagbox}
\usepackage{multirow}
\usepackage{pdfpages}

\makeatletter
\newcommand{\printfnsymbol}[1]{%
  \textsuperscript{\@fnsymbol{#1}}%
}
\makeatother
\cvprfinalcopy 

\def\cvprPaperID{2967} 

\ifcvprfinal\fi
\begin{document}

\title{Estimating 6D Pose From Localizing Designated Surface Keypoints}

\author{Zelin Zhao\\
{\tt\small sjtuytc@sjtu.edu.cn}
\and
Gao Peng\thanks{contribute equally}\\
{\tt\small ecr23pg@gmail.com}
\and
Haoyu Wang\printfnsymbol{1}\\
{\tt\small why2011btv@sjtu.edu.cn}
\and
Hao-Shu Fang\\
{\tt\small fhaoshu@gmail.com}
\and
Chengkun Li\\
{\tt\small sjtulck@sjtu.edu.cn}
\and
Cewu Lu\thanks{corresponding author} \\
{\tt\small lucewu@sjtu.edu.cn}
}

\maketitle
\newcommand{\cewu}[1]{\textcolor{red}{$_{cewu}$[#1]}}
\newcommand{\haoshu}[1]{\textcolor{red}{$_{haoshu}$[#1]}}
\newcommand{\haoyu}[1]{\textcolor{red}{$_{haoyu}$[#1]
}}

\begin{abstract}
In this paper, we present an accurate yet effective solution for 6D pose estimation from an RGB image. The core of our approach is that we first designate a set of surface points on target object model as keypoints and then train a keypoint detector (KPD) to localize them. Finally a PnP algorithm can recover the 6D pose according to the 2D-3D relationship of keypoints. Different from recent state-of-the-art CNN-based approaches \cite{Kehl2017SSD6DMR, Rad2017BB8AS} that rely on a time-consuming post-processing procedure, our method can achieve competitive accuracy without any refinement after pose prediction. Meanwhile, we obtain a 30\% relative improvement in terms of ADD accuracy \cite{Hinterstoisser:2012:MBT:2481913.2481959} among methods without using refinement. Moreover, we succeed in handling heavy occlusion by selecting the most confident keypoints to recover the 6D pose. For the sake of reproducibility, we will make our code and models publicly available soon.
\end{abstract}

\section{Introduction}
The task of 6D pose estimation is now grabbing more and more attention because of its significance in robotics, virtual reality and augmented reality. Recently, the CNN-based methods \cite{Kehl2017SSD6DMR, Rad2017BB8AS, tekin18, 7298758, xiang2018posecnn} are providing terrific results in 6D pose estimation area without relying on rich texture information. Since the depth cameras are vulnerable in the wild or on specular objects and they consume too much power being an active sensor on mobile devices, some recent literature \cite{Kehl2017SSD6DMR, Rad2017BB8AS, tekin18} proposes to estimate 6D pose directly from a single RGB image without using depth information. Methods proposed in \cite{Kehl2017SSD6DMR, Rad2017BB8AS} can get convincing results which are competitive with those leveraging RGB-D data. Yet they both rely on an effective but time-consuming refinement procedure in order to achieve accurate pose estimation. \cite{tekin18} proposes to eliminate the refinement to obtain a huge speed gain. However, they yield worse accuracy results in terms of ADD metric \cite{Hinterstoisser:2012:MBT:2481913.2481959}.

In this paper we argue for a fast and accurate approach. We reckon that recovering 6D pose from estimating surface keypoints is more natural and easier than from predicting the viewpoints because the surface keypoints are more directly and closely associated with features of the target object. Another benefit of predicting a number of surface keypoints is the robustness against occlusion since even if some keypoints are invisible due to partial occlusion, three remaining ones are sufficient to estimating 6D pose by Perspective-n-point (PnP) algorithm \cite{Lepetit2008EPnPAA}. Therefore we propose to predict 6D pose from localizing surface keypoints.

We first select $k$ 3D keypoints in the model offline by 3D SIFT algorithm \cite{Scovanner:2007:SDA:1291233.1291311} and generate a dataset with bounding boxes and keypoints annotated. After that, we separately train the object detector YOLOv3 \cite{DBLP:journals/corr/abs-1804-02767} and a keypoint detector (KPD). As shown in human pose estimation literature, localizing keypoints by predicting heatmaps \cite{Newell2016StackedHN, tompson2014joint, wei2016convolutional} is better at regressing keypoints coordinates directly \cite{toshev2014deeppose}. Therefore in our KPD we adopt a heatmap localization strategy to estimate the locations of keypoints.

\begin{figure*}[t]
\begin{center}
   \includegraphics[width=\linewidth]{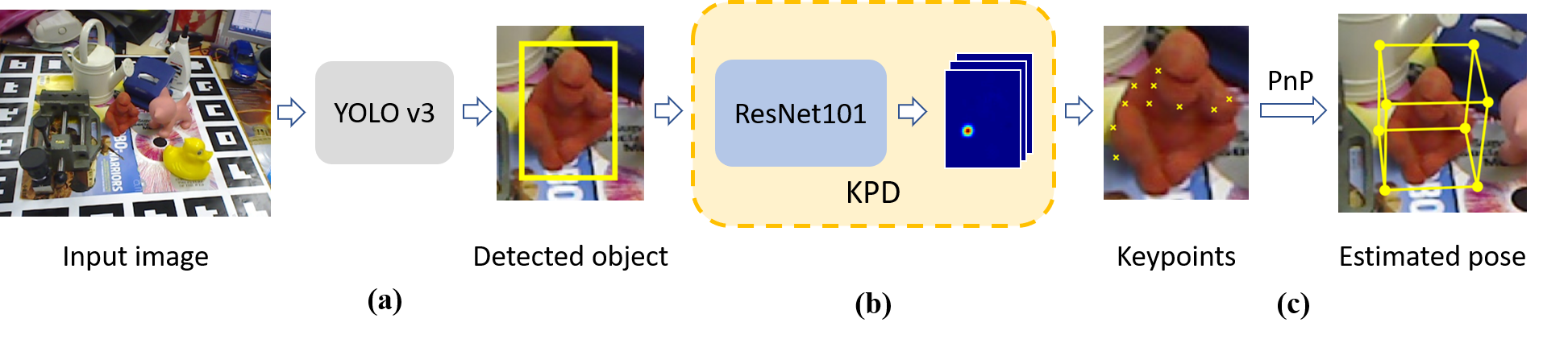}
\end{center}
   \caption{Visualization of our proposed pipeline. We first (a) detect bounding box then (b) localize the designated keypoints using keypoint detector (KPD). Finally (c) we use a PnP algorithm to recover the 6D pose.}
\label{fig:pipeline}
\end{figure*}

During testing stage, the target object is detected by YOLOv3 and then its 2D keypoints are localized by the KPD. Thanks to the Perspective-n-point (PnP) \cite{Lepetit2008EPnPAA}, we can estimate the object's 6d pose via the 2D-3D correspondences of keypoints. In order to deal with partial occlusion, we propose to select the most confident keypoints before running PnP algorithm. Figure \ref{fig:pipeline} illustrates our proposed pipeline.

We evaluate our architecture on two benchmark datasets: LineMod \cite{Hinterstoisser:2012:MBT:2481913.2481959} and Occlusion \cite{10.1007/978-3-319-10605-2_35}. Concerning accuracy, we not only outperform the state-of-the-art not-using-refinement method \cite{tekin18} by a large margin (30\% relatively) but also achieve competitive results with using-refinement methods. Meanwhile, we are much faster than using-refinement techniques due to the elimination of the time-consuming refinement procedure.


Our key contribution is that we present a novel method of 6D pose estimation based on localizing designated surface keypoints. We show the great potential of RGB-only methods without refinement by achieving state-of-the-art results. Moreover, we conduct several ablation studies to verify the effectiveness of key components in our architecture.

\section{Related Work}
There is a large number of literature on 6D pose estimation. We first review them from different perspectives and then have a short glance at human pose estimation work where we get inspiration.

\paragraph{Classical Feature Mapping Methods} Many classical approaches aim to find suitable feature descriptors and predicting pose by feature matching. Scale-invariant approaches\cite{10.1007/978-3-319-10605-2_35, 7780759, Kehl2016DeepLO, tejani2014latent} use depth information to extract or learn features which are robust to different lighting conditions and even partial occlusions. Other approaches \cite{5540108, Yi2016LIFTLI, 10.1007/978-3-642-15558-1_26} attempt to find local keypoints and then vote for the orientations. They either require certain object textural property or are not robust enough for handling occlusion.

\paragraph{CNN-based Methods} Deep learning in 6D pose estimation have demonstrated its great power and potential. \cite{7298758} points out that one can recover 6D pose from figures by estimating viewpoints or keypoints. The viewpoint based method  \cite{Kehl2017SSD6DMR} extends SSD \cite{10.1007/978-3-319-46448-0_2} and constructs an end-to-end architecture outputting viewpoints and inplane-rotations. Then it uses the projection relationship to lift the object and further refine estimated poses. \cite{Sundermeyer_2018_ECCV} creatively learns implicit orientation by Augmented Autoencoder. The keypoints based methods \cite{Rad2017BB8AS, tekin18} attempt to predict corner points of the bounding boxes and then use PnP algorithm to estimate 6D pose. \cite{pavlakos17object3d} estimates locations of semantic keypoints instead. Other methods \cite{Kendall2015PoseNetAC, xiang2018posecnn} make an effort to train translation and rotation predictors then combine them to get the 6D pose.

\paragraph{RGB Only Without Refinement} Since it is not always possible to use the depth cameras which consume much power on mobile devices and are easy to fail in the open air, methods only using RGB images to recover 6D pose have been demonstrated to be competitive \cite{Kehl2017SSD6DMR, Mousavian20173DBB, Poirson2016FastSS} with a natural advantage of cheap data requirement. Furthermore, our baseline \cite{tekin18} eliminates the refinement procedure which is time-consuming and thus not suitable for real-time applications such as virtual reality and robots grasping. Moreover, the Iterative Closest Points (ICP) refinement needs tricky tuning procedures to work well, especially when only RGB images are given. Therefore, we also take an RGB-only method and not use the post-processing procedure.

\paragraph{Person Pose Estimation} Deep learning methods have dominated person pose estimation tasks in recent years, not only in single person pose estimation \cite{Jain2013LearningHP, Ouyang2014MultisourceDL, 6909610} but also in multiple person pose estimation \cite{cao2017realtime, fang2017rmpe, He2017MaskR}. \cite{fang2017rmpe} proposes a two-step accurate and fast framework for multiple pose estimation. Given input images, they first use a human detector to draw bounding boxes and then utilize a single person pose estimator (SPPE) to predict keypoints. We are inspired by them and adopt their two-step techniques followed by the PnP algorithm to get the final 6D pose.

\section{Methodology}
We can transform the 6D pose estimation problem into detecting 2D image coordinates of keypoints whose 3D coordinates in model space are previously known by us. Given the 2D coordinate predictions and their associated 3D coordinates in model space, we can recover 6D pose via a Perspective-n-point (PnP) \cite{Lepetit2008EPnPAA}. Different from \cite{Rad2017BB8AS, tekin18}, our keypoints are designated by us on the model surface via a 3D SIFT \cite{Scovanner:2007:SDA:1291233.1291311} algorithm. We now present our pipeline and describe the critical procedures in detail.
\subsection{Our Pipeline}
\label{sec:pipeline}
As shown in Figure \ref{fig:pipeline}, the input to our architecture is an RGB image $I$. Our goal is to estimate 6D pose $\mathbf{P} = \big[\mathbf{R}\big|\mathbf{t}\big]$ of the target object, where $\mathbf{R}$ is the rotation matrix and $\mathbf{t}$ is the translation matrix.

Utilizing the accurate and efficient detector YOLOv3 \cite{DBLP:journals/corr/abs-1804-02767}, our first stage aims at detecting bounding box of the target object. Then the image $I$ is cropped by this bounding box and then resized to a new image $I^{*}$ with only one proposed object. Then $I^{*}$ is sent to the keypoint detector (KPD).

In the second stage, the KPD is designed for localizing 2D keypoints which are previously designated in 3D model space. (Keypoint designation will be discussed in Section \ref{sec:Synthetic_dataset}.) We use ResNet-101 \cite{DBLP:journals/corr/HeZRS15} as the backbone of Keypoint Detector. The KPD takes $I^{*}$ as input and outputs $k$ heatmaps corresponding to $k$ 2D points. For the $p$-th keypoint, we consider the pixel with the maximum value in the $p$-th heatmap as its estimated location. That is,
\begin{equation}
\label{eq:max_in_heatmap}
    \mathcal{H}_{x,y}^{p} = \max_{i}(\max_{j} \mathcal{H}_{i,j}^{p})
\end{equation}
where $(x,y)$ is the estimated coordinates of the $p$-th keypoint in $I^{*}$. We futher denote $\mathcal{H}_{x,y}^{p}$ as the confidence of the $p$-th estimated keypoint. With detected bounding box, we can derive the estimated location of this $p$-th keypoint in the original image $I$. In this way, we store 2D coordinates of all keypoints in matrix $M_{k2D}$.

In cases where the objects are under occlusion, the keypoints are ranked by their confidence values and only the $l$ most confident keypoints are preserved and utilized in the third stage because they are normally more accurate. Theoretically, $l\geqslant 3$ is sufficient to estimate 3D pose while $l$ can be larger than three in practice to boost the predicting precision. Our further experiment (See Section \ref{sec:choosing_most_conf}) shows that this selecting procedure is effective when handling occlusion.

During the third stage, we use the Perspective-n-point (PnP) \cite{Lepetit2008EPnPAA} algorithm to recover 6D pose via the relationship between 2D and 3D keypoints. Let $M_{k2D}$ and $M_{k3D}$ be the 2D and 3D coordintes of keypoints in image space and model space respectively, the transformation relationship is:
\begin{equation}
\label{eq:transformation}
    M_{k2D} = K\big[\mathbf{R}\big|\mathbf{t}\big]M_{k3D}
\end{equation}
where both $M_{k3D}$ and $M_{k2D}$ are represented in homogeneous form, and $K$ is the known camera intrinsic matrix. Given $M_{k3D}$ and $M_{k2D}$, the PnP algorithm can figure out the rotation matrix $\mathbf{R}$ and translation matrix $\mathbf{t}$ through the rigid relationship of target object. Mathmatically, the PnP algorithm solve the matrix $[\mathbf{R}\big|\mathbf{t}\big]$ by simple least square optimaztion.

\begin{figure*}[t]
\begin{center}
   \includegraphics[width=0.8\linewidth]{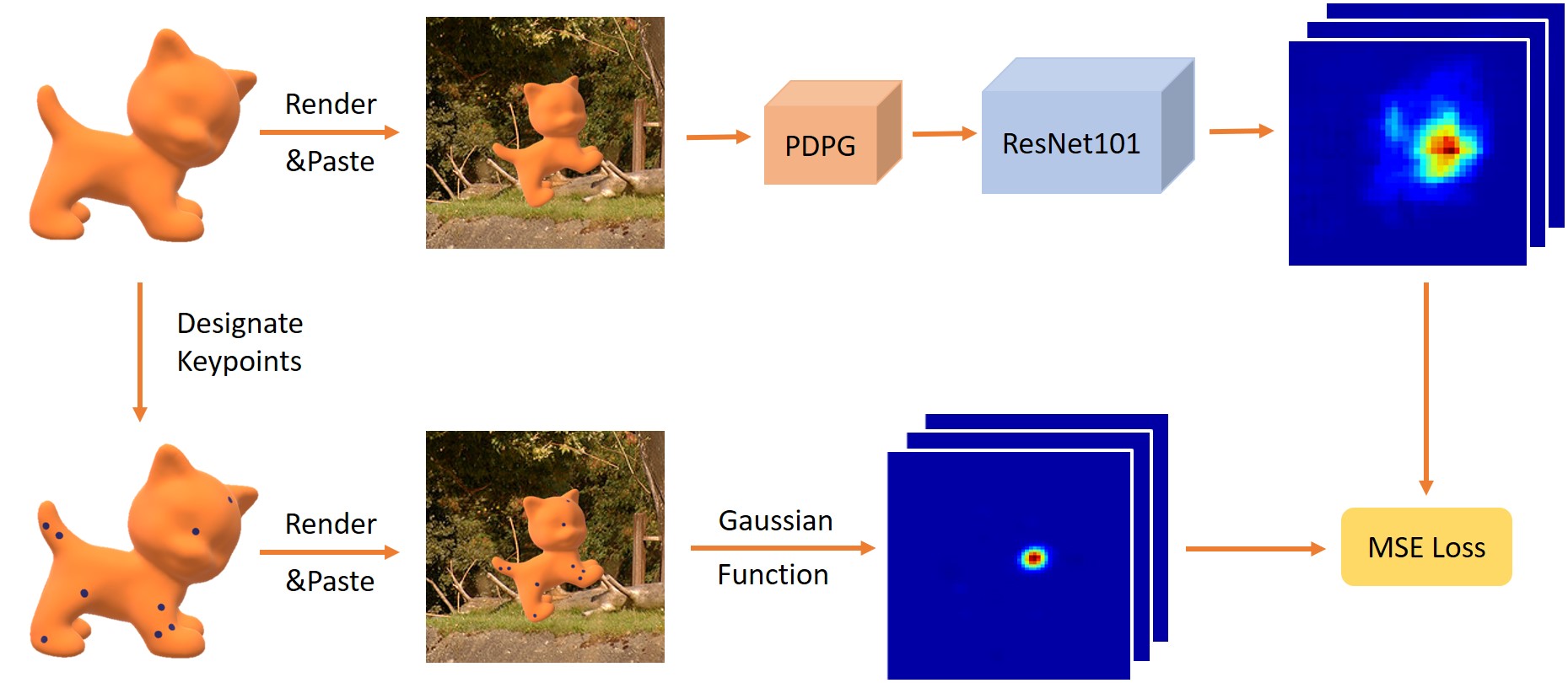}
\end{center}
   \caption{Generating dataset with keypoints annotations and training KPD to localize them. Upper row: We render the model and paste it onto RGB image. PGPG can augment input data dealing with inaccurate bounding box \cite{fang2017rmpe}. The followed ResNet101 is trained to generate heatmaps corresponding to $k$ designated keypoints. Lower row: We designate keypoints from 3D model and transfer them into RGB images via equation \ref{eq:transformation}. Then $k$ ground truth heatmaps are generated according to the 2D locations of $k$ keypoints via a Gaussian Function. We train the KPD by minimizing the MSE Loss between predicted and ground truth heatmaps. More examples of designated keypoints can be found in supplementary.}
\label{fig:training_kpd}
\end{figure*}
\subsection{Keypoint Designation and Annotation}
\label{sec:Synthetic_dataset}

Formally, let $M_{3D}$ be the point cloud of the object model, keypoint designation means that we select a set of points in model space as the keypoints $M_{k3D}$. In our proposed pipeline, neither the number nor the type of keypoints is restricted, but it would be better to designate keypoints with strong representative ability. We reckon that surface points are a better choice than the vertices alongside bounding boxes because the surface points are closely related with the model features. Since 3D Scale-invariant feature transform (SIFT) algorithm \cite{Scovanner:2007:SDA:1291233.1291311} can extract local features which are invariant to rotation, scale and robust to different illumination conditions, we adopt SIFT keypoints as our designated keypoints.

We need to annotate those designated keypoints in images for training keypoint detector (KPD). The designated 3D keypoints are projected to image space applying Equation \ref{eq:transformation}, deriving the ground truth coordinates of $k$ keypoints in image space. During generating synthetic images via rendering and pasting as used in \cite{Kehl2017SSD6DMR,Sundermeyer_2018_ECCV,Rad2017BB8AS}, we can obtain the virtual camera intrinsic matrix $K$ and assigned rotation-translation matrix $\big[\mathbf{R}\big|\mathbf{t}\big]$ which are needed by Equation \ref{eq:transformation}. As for the dataset with real pose annotated, $K$, $\mathbf{R}$ and $\mathbf{t}$ are given directly.

Since human pose estimation literature has shown that localizing keypoints by predicting heatmaps \cite{Newell2016StackedHN, tompson2014joint, wei2016convolutional} is better at regressing keypoints coordinates directly \cite{toshev2014deeppose}, we employ the Gaussian function on the 2D coordinates and generate $k$ ground truth heatmaps for the $k$ keypoints. Formally, let $\mathbf{x}^{p}$ be the ground truth location of $p$-th keypoint and its corresponding heatmap is $\mathcal{H}^{p}$. The value at location $\mathbf{p} \in \mathbb{R}^{2}$ in $\mathcal{H}^{p}$ is defined as
\begin{equation}
\label{eq:gaussian}
\mathcal{H}^{p}( \mathbf { p } ) = \exp \left( - \frac { \left\| \mathbf { p } - \mathbf{x}^{p} \right\| ^ { 2 } } { \sigma ^ { 2 } } \right)
\end{equation}
where $\sigma$ is a parameter controlling the spread of the peak. The ground truth heatmaps are used for training KPD.

\subsection{Training and Testing}
Different from the end-to-end training in \cite{tekin18}, we train the YOLO detector and the KPD separately. Training in two stages can help decouple the problem of object detection and keypoint localization so that we can address them one by one, leading to better overall results.

Feeding our annotated dataset, we train YOLOv3 using common settings as described in \cite{DBLP:journals/corr/abs-1804-02767} and KPD by minimizing the MSE loss between the ground truth and predicted heatmaps. Moreover, to deal with the bias in bounding box detection which would draw to an error in keypoint localization, we use the developed Pose-guided Proposal Generator (PGPG) proposed by \cite{fang2017rmpe} to augment the training data. The whole procedure of generating dataset and training KPD is shown in Figure \ref{fig:training_kpd}.

During testing, we load the trained weights of both YOLO and KPD to the overall architecture. As described in Section \ref{sec:pipeline}, our pipeline first detects the bounding box, then localizes the keypoints. Finally, the PnP algorithm will help recover the 6D pose.

\begin{figure}[t]
\begin{center}
   \includegraphics[width=0.9\linewidth]{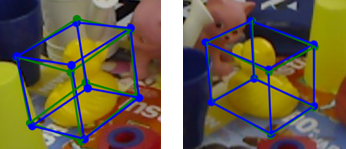}
   \includegraphics[width=0.9\linewidth]{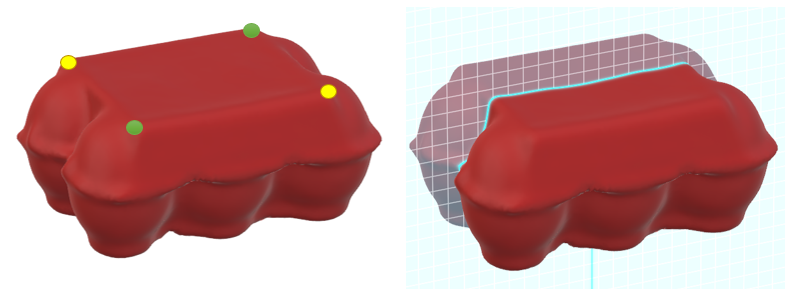}
\end{center}
   \caption{Dealing with symmetry. Top: Our pipeline can handle two ducks in mirror symmetry without special care. Lower Left: the $\pi$-rotational ($\alpha = \pi$) symmetric object eggbox has pairs of indistinguishable keypoints in 3D space. (One pair in green and the other in yellow.) Lower Right: We can handle the $\pi$-rotational symmetry by restricting training images in half of the 3D space, just as if we only look at the eggbox from the front.}
\label{fig:symmetry_explain}
\end{figure}
\subsection{Handle Symmetry Cases}

Estimating a keypoint $p_0$ located on a symmetric object may run into problem for there may be several points appearing the same as $p_0$. We need to handle different types of symmetry carefully (See Figure \ref{fig:symmetry_explain}).

\paragraph{Mirror Symmetry} A pair of points with mirror symmetry are in fact distinguishable for one of them is on the left and the other is on the right. Our KPD can learn to distinguish the left side from the right natually from end-to-end training and thus the mirror symmetry can be handled without special care which is verified by our experiments.

\paragraph{$\alpha$-rotational Symmetry} We consider that an object has $\alpha$-rotational symmetry if it looks the same after $\alpha$ angle rotation. For a designated keypoint $p$ on an object with $\alpha$-rotation symmetry, there are one or more points in symmetry with $p$ and they can be mistaken as $p$, resulting in divergence in training. To address such ambiguity, the training images should be projected from the 3D space that is within the range $[0,\alpha]$ in cylindrical coordinates system around the symmetric axis (see lower part of Figure \ref{fig:symmetry_explain}). On the other side, as used in \cite{7780735,Hinterstoisser:2012:MBT:2481913.2481959,tekin18}, we take symmetry into account when evaluating the results. Specifically, our predicted pose of an  $\alpha$-rotational object may vary due to the rotational symmetry, but the various poses should all be considered as correct as long as they are symmetric with ground truth pose.

\section{Experiments}

We evaluate our proposed pipeline in two challenging benchmark datasets: LineMod \cite{Hinterstoisser:2012:MBT:2481913.2481959} and Occlusion \cite{10.1007/978-3-319-10605-2_35}. We first describe two datasets and evaluation metrics in short and give the implementation details. Then both qualitative and quantitative results are presented.

\begin{table*}[t]

\centering
\setlength{\tabcolsep}{1.3mm}{
\begin{tabular}{ll|c|c|c|c|c|c|c|c|c|c|c|c|c|c}
                                              &           & \multicolumn{1}{l|}{Ape} & \multicolumn{1}{l|}{Bvise} & \multicolumn{1}{l|}{Cam} & \multicolumn{1}{l|}{Can} & \multicolumn{1}{l|}{Cat} & \multicolumn{1}{l|}{Drill} & \multicolumn{1}{l|}{Duck} & \multicolumn{1}{l|}{Box} & \multicolumn{1}{l|}{Glue} & \multicolumn{1}{l|}{Holep} & \multicolumn{1}{l|}{Iron} & \multicolumn{1}{l|}{Lamp} & \multicolumn{1}{l|}{Phone} & \multicolumn{1}{l}{Avg} \\ \hline
\multicolumn{1}{c|}{\multirow{4}{*}{w/o}}     & BB8\cite{Rad2017BB8AS}       & 27.9                     & 62.0                           & 40.1                     & 48.1                     & 45.2                     & 58.6                         & 32.8                      & 40.0                        & 27.0                      & 42.4                             & 67.0                      & 39.9                      & 35.2                       & 43.6                        \\
\multicolumn{1}{c|}{}  & SSD-6D \cite{Kehl2017SSD6DMR}    & 0                        & 0.2                           & 0.4                     & 1.4                     & 0.5                     & 2.6                         & 0                         & 8.9       & 0    & 0.3     & 8.9  & 8.2  & 0.2  & 2.4                        \\
\multicolumn{1}{c|}{}                         & Tekin \cite{tekin18}     & 21.6                     & 81.8                           & 36.6                     & 68.8                     & 41.8                     & 63.5                         & 27.2                      & 69.6                        & \textit{\textbf{80.0}}                      & 42.6                             & 75.0                     & 71.1                     & 47.7                      & 56.0                       \\
\multicolumn{1}{c|}{}                         & OURS      & \textit{\textbf{41.2} }                    & \textit{\textbf{85.7} }                          & \textbf{78.9}                     & \textit{\textbf{85.2}}                     & \textbf{73.9}                     & \textbf{77.0}                         & \textit{\textbf{42.7}}                      & \textit{\textbf{78.9}}                        & 72.5                      & \textit{\textbf{63.9}}                             & \textbf{94.4}                      & \textbf{98.1}                      & \textit{\textbf{51.0}}                       & \textbf{\textit{72.6}}                        \\ \hline
\multicolumn{1}{l|}{\multirow{3}{*}{w/Ref.}} & Brachmann \cite{7780735}& 33.2                     & 64.8                           & 38.4                     & 62.9                     & 42.7                     & 61.9                         & 30.2                      & 49.9                        & 31.2                      & 52.8                             & 80.0                      & 67.0                      & 38.1                       & 50.2                        \\
\multicolumn{1}{l|}{}                         & BB8 \cite{Rad2017BB8AS}       & 40.4                     & \textbf{91.8}                           & 55.7                     & 64.1                     & 62.6                     & 74.4                         & 44.3                      & 57.8                        & 41.2                      & \textbf{67.2}                             & 84.7                      & 76.5                      & 54.0                       & 62.7                        \\
\multicolumn{1}{l|}{}                         & SSD-6D \cite{Kehl2017SSD6DMR}    & \textbf{65}                       & 80                             & 78                       & \textbf{86}                       & 70                       & 73                           & \textbf{66}                        & \textbf{100}                         & \textbf{100}                       & 49                               & 78    & 73 & \textbf{79}                         & \textbf{79}
\end{tabular}}
\caption{Accuracy comparison of methods with refinement or without refinement in terms of ADD metric on the LineMod dataset. The overall best numbers are represented in \textbf{bold} and the best numbers in methods without refinement are represented in \textbf{\textit{bold and italic}}.}
\label{tb:ADD}
\end{table*}
\subsection{Benchmark Datasets}
\paragraph{LineMod \cite{Hinterstoisser:2012:MBT:2481913.2481959}} is a standard benchmark dataset for 6D pose estimation. The LineMod dataset contains 18273 test images for 15 objects. The central object in the RGB image is considered as the target object whose bounding box, rotation and translation matrices are annotated. The 3D models with surface color are also provided and thus we can generate synthetic images for training. As others, we will skip the third and seventh objects which lack a meshed model.
\paragraph{Occlusion \cite{10.1007/978-3-319-10605-2_35}}
is a dataset for multi-object detection and 6D pose estimation. It is created from LineMod dataset by denoting extra bounding boxes and poses of all the seven kinds of other objects appearing in the Benchvise sequence. As its name implies, most objects in the Occlusion dataset are under partial occlusion, making the multi-object detection and pose recovery quite difficult.

\subsection{Evaluation Metrics}
We use two metrics to evaluate the 6D pose estimation results. The \textbf{2D reprojection error} \cite{7780735} is the mean distance between the 2D projection of the object's 3D mesh vertices applying the predicted and the ground truth pose, and the predicted pose is correct if the error is less than 5 pixels. It mainly measures if the estimated pose is visually acceptable, so they are suitable for applications in virtual or augmented reality. The \textbf{ADD error} \cite{Hinterstoisser:2012:MBT:2481913.2481959} is the mean 3D distance between model vertices transformed by the ground truth pose and by the predicted pose. Formally,
\begin{equation}
\label{eq:add_error}
    \Delta_{ADD} = \frac { 1 } { |M | } \sum _ { x \in  M } \left\| ( \mathbf { R } \mathbf { x } + \mathbf { t } ) - ( \hat { \mathbf { R } }  \mathbf {x} + \hat { \mathbf { t } }) \right\|
\end{equation}
where $M$ is the set of model vertices, $\mathbf{R}$ and $\mathbf{t}$ are the predicted rotation and translation matrices while $\hat { \mathbf { R }}$ and $\hat { \mathbf { t } }$ are the ground truth ones. The estimated pose is considered to be correct if the ADD error is smaller than 10\% of the object's diameter. The ADD metric is stricter because it requires the predicted pose to be accurate enough to limit the ADD error within around $1 \sim 2$ cm. We use the terminologies 2D reprojection accuracy and ADD accuracy to represent the percentage of correct poses among all predicted poses using the above two metrics correspondingly.

 For evaluating the symmetric objects, the 2D reprojection error can naturally accept symmetric poses as long as the objects with pose applied are visualized the same in image space. However, the ADD metric must be changed slightly to deal with symmetry cases. As in \cite{7780735,Hinterstoisser:2012:MBT:2481913.2481959,tekin18}, we change Equation \ref{eq:add_error}  to
\begin{equation}
\label{eq:sym_add_error}
    \Delta_{ADD}' = \frac { 1 } { |M | } \sum _ { x_1 \in  M } \min_{ x_2 \in M}\left\| ( \mathbf { R } \mathbf { x_1 } + \mathbf { t } ) - ( \hat { \mathbf { R } }  \mathbf {x_2} + \hat { \mathbf { t } }) \right\|
\end{equation}
which is looser than the normal form. We only use this form when the object has rotational symmetry such as the eggbox in LineMod.

\subsection{Implementation Details}
The overall architecture is implemented with PyTroch 0.4.1 and run on an i9-7900X CPU @3.30GHz with an NVIDIA Geforce GTX 1080Ti.

\paragraph{Generating Training Dataset} Besides the training images provided by \cite{tekin18}, we also render about 30000 synthetic images using OpenGL and annotate keypoints on them for training as aforementioned in Section \ref{sec:Synthetic_dataset}. We tune the parameters of SIFT to set the total number of designated keypoints to be 50 for all objects except for the eggbox where we choose 17 unsymmetric keypoints.

\paragraph{Training Setting} For the first stage, we train YOLO using stochastic gradient descent for optimization, with learning rate initially set as 0.001 and divided by 10 at every 100 epochs. We randomly change the hue, saturation and exposure of images by up to a factor of 1.5 while we also randomly scale the image by up to a factor of 20\% of the image size. For the second stage, we train the KPD using the Adam optimizer \cite{Kingma2014AdamAM} and the learning rate is set as a constant 0.001. To increase the inference speed, we use multi-process during training and testing.

\subsection{Accuracy and Speed on LineMod Dataset}

2D reprojection accuracy results are shown in table \ref{tb:reprojection_accuracy}. Our method outperforms other state-of-the-art methods. The state-of-the-art not-using-refinement method \cite{tekin18} has achieved a slightly better results than the using-refinement methods \cite{7780735, Rad2017BB8AS}. We take a step forward and show the advantage of not-using-refinement methods in terms of 2D reprojection accuracy.

ADD accuracy results are shown in table \ref{tb:ADD}. We not only overtake the state-of-the-art not-using-refinement methods by a large margin (30\% relatively) but also break the domination of using-refinement methods by achieving competitive results. We can see those using-refinement methods work poorly if the refinement procedure is eliminated. For objects which have rich information in their outline (such as Benchvise, Iron and Lamp) or objects whose surface curvature varies greatly (such as Cat and Camera), our method performs very well. Nevertheless, our method is not that satisfactory when dealing with symmetric objects (such as eggbox) or objects which have little surface information (such as phone and glue) in comparison with \cite{Kehl2017SSD6DMR}, the state-of-the-art using-refinement method. In general, we beat \cite{Kehl2017SSD6DMR} at 6 out of 13 sequences, showing the great potential of not-using-refinement methods. Some qualitative results can be found in Figure \ref{fig:visualize_all}.

\begin{table}[t]
\begin{center}
\begin{tabular}{|c|c|c|}
\hline
                        & Method    & Average       \\ \hline
\multirow{3}{*}{w/o}    & BB8 \cite{Rad2017BB8AS}       & 83.9          \\ \cline{2-3}
                        & Tekin \cite{tekin18} & 90.4          \\ \cline{2-3}
                        & OURS      & \textbf{94.5} \\ \hline
\multirow{2}{*}{w/Ref.} & Brachmann \cite{7780735} & 73.7          \\ \cline{2-3}
                        & BB8 \cite{Rad2017BB8AS}       & 89.3          \\ \hline

\end{tabular}
\end{center}
\caption{Comparison of 2D reprojection Accuracy on LineMod. The pixel threshold is 5.}
\label{tb:reprojection_accuracy}
\end{table}

Our speed results are shown in table \ref{tb:speed}. We can see the post-refinement procedures are time-consuming. Since we can achieve state-of-the-art accuracy even without refinement, we can feel free to eliminate the refinement procedure. \cite{tekin18} is faster than us because they use a one-shot design.
\begin{table}[]
\begin{center}
\begin{tabular}{c|c|c}
\hline
Method     & Speed (fps) & Ref. Time (ms)\\ \hline
Branchmann \cite{7780735} & 2  &100                 \\
BB8 \cite{Rad2017BB8AS}  & 3  &21                 \\
SSD-6D \cite{Kehl2017SSD6DMR}    & 10  &24                \\
Tekin \cite{tekin18}     & \textbf{45} &-     \\
OURS       & 25  &-               \\ \hline
\end{tabular}
\end{center}
\caption{Speed results of state-of-the-art methods. \cite{tekin18} and OURS don't use the refinement procedure. (We retest the speed of \cite{tekin18} using their published codes on our server for a fair comparison.)}
\label{tb:speed}
\end{table}

\subsection{Results on the Occlusion Dataset}
\label{sec:occlusionresult}
\begin{figure}[t]
\begin{center}
   \includegraphics[width=\linewidth]{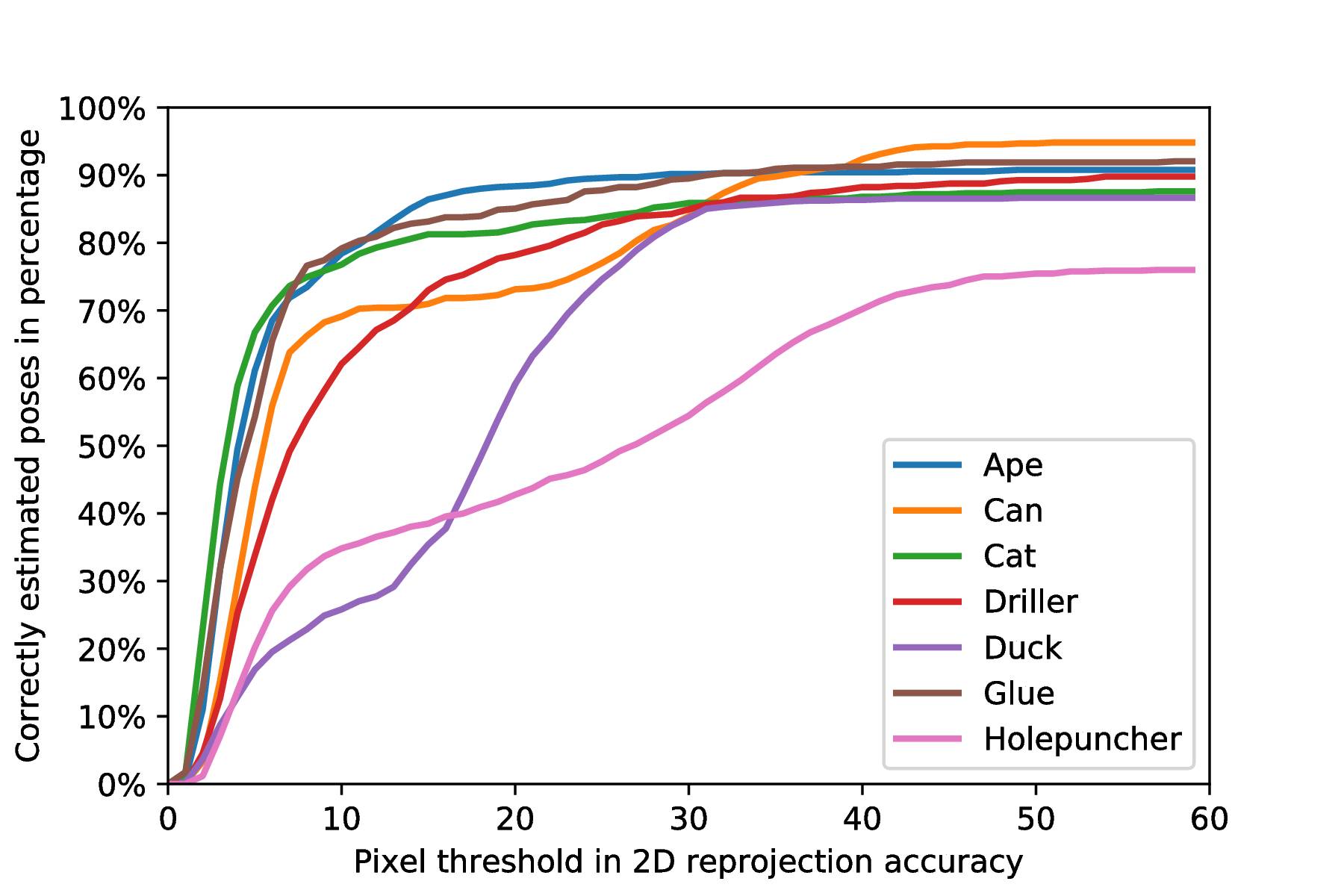}
    \includegraphics[width=0.9\linewidth]{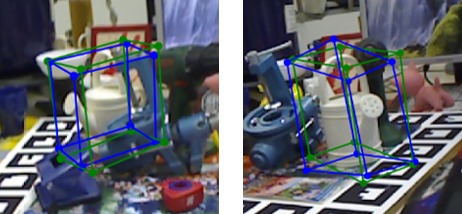}
\end{center}
   \caption{Results on the Occlusion dataset. Top: Percentage of correctly estimated poses as a function of the pixel threshold in 2D reprojection accuracy. We can correctly estimate about 65\% frames for a 15px threshold and about 80 \% frames for a 30 px threshold on average. Bottom: Two frames with a 15px and 30px error respectively. The green and blue rendered bounding boxes denote the ground truth and predicted pose respectively.}
\label{fig:occlusion_result}
\end{figure}
\begin{figure*}[t]
\begin{center}
   \includegraphics[width=0.90\linewidth]{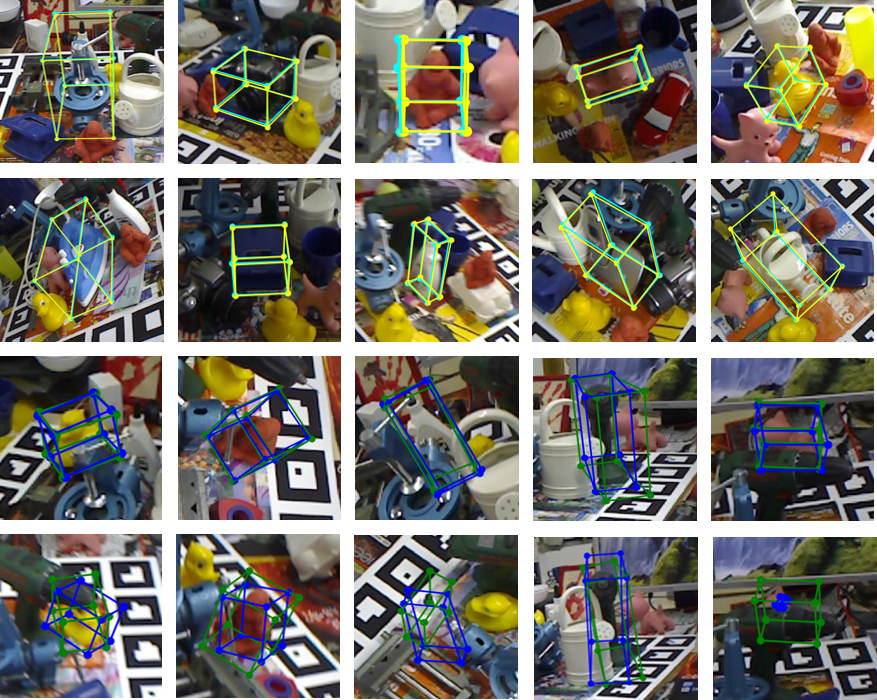}
\end{center}
   \caption{Qualitative results on LineMod and Occlusion datasets. First two rows: Results on the LineMod dataset, our method can estimate 6D pose correctly in challenging scenes with extreme lighting conditions, heavy clutter and motion blur. Third row: Result on the Occlusion dataset. We can still recover pose correctly when partial occlusion exists. Last row: Failure cases on the Occlusion dataset due to severe blur or overly deficient feature information.}
\label{fig:visualize_all}
\end{figure*}

For experiments on the LineMod dataset, we use all the predicted keypoints to estimate 6D pose and do not abandon any predicted keypoints. But when on the Occlusion dataset we only keep $l=10$ most confident keypoints and abandon other low-confident keypoints. The results are shown in Figure \ref{fig:occlusion_result}. The state-of-the-art not-using-refinement method \cite{tekin18} chooses 183 images in the Occlusion dataset to train and others to test so there are very similar occlusions between the training and testing images. We follow the strict experiment condition with \cite{Rad2017BB8AS} that we do not use any images in the test sequence. Although the training process is more difficult, we still achieve much better results than  \cite{tekin18}. For a 15px threshold in 2D reprojection accuracy, \cite{tekin18} can estimate about 50\% frames correctly while we can estimate around 65\% frames correctly, closer to 80\% in state-of-the-art using-refinement method \cite{Rad2017BB8AS}. Some qualitative results are in Figure \ref{fig:visualize_all}.

\section{Analysis and Ablation Study}

\subsection{Inferring Invisible Back-face Keypoints}
\begin{figure}[t]
\begin{center}
   \includegraphics[width=\linewidth]{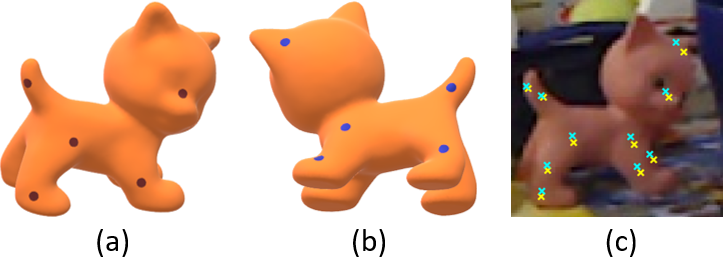}
\end{center}
   \caption{Infering back-face points. Our pipeline can predict keypoints not only on the front (a) but also on the back (b) of the model. The keypoints are all estimated precisely without distinction. (c)}
\label{fig:infer_backpoints}
\end{figure}

When estimating surface keypoints from a single RGB image, some keypoints can be invisible because they are on the back of the 3D object. However, this seemingly frustrating problem can be handled naturally by our architecture. Since the investigated objects are rigid with 6 degrees of freedom (DoF), our CNN can learn to infer back-face keypoints from end-to-end training without special treatment. In other words, although only a few keypoints on the front of the 3D object can be seen directly from the RGB image, our trained pipeline can still predict all the designated keypoints including invisible ones at high enough accuracy (See Figure \ref{fig:infer_backpoints}). This portrait is worth noticing and very beneficial for estimating 6D pose from localizing surface keypoints.

\subsection{Selecting Most Confident Keypoints}
\label{sec:choosing_most_conf}
\begin{figure}[t]
\begin{center}
   \includegraphics[width=0.85\linewidth]{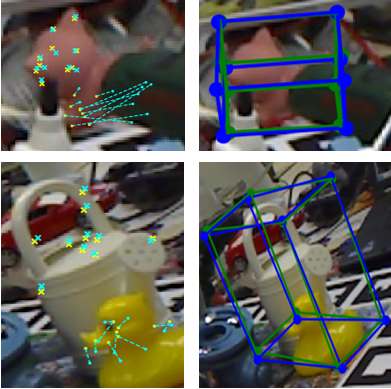}
\end{center}
   \caption{Visualize the selecting procedure. Left: The ground truth keypoints (yellow) and predicted keypoints (blue). The top-10 confident points are plotted with crosses while others are randomly sampled and plotted with dots. We can see that the top-10 confident points are predicted accurately, but those unselected unconfident keypoints are often biased due to occlusion. Right: Ground truth poses (green) and predicted poses (blue).}
\label{fig:filter10}
\end{figure}

\begin{figure}[t]
\begin{center}
   \includegraphics[width=\linewidth]{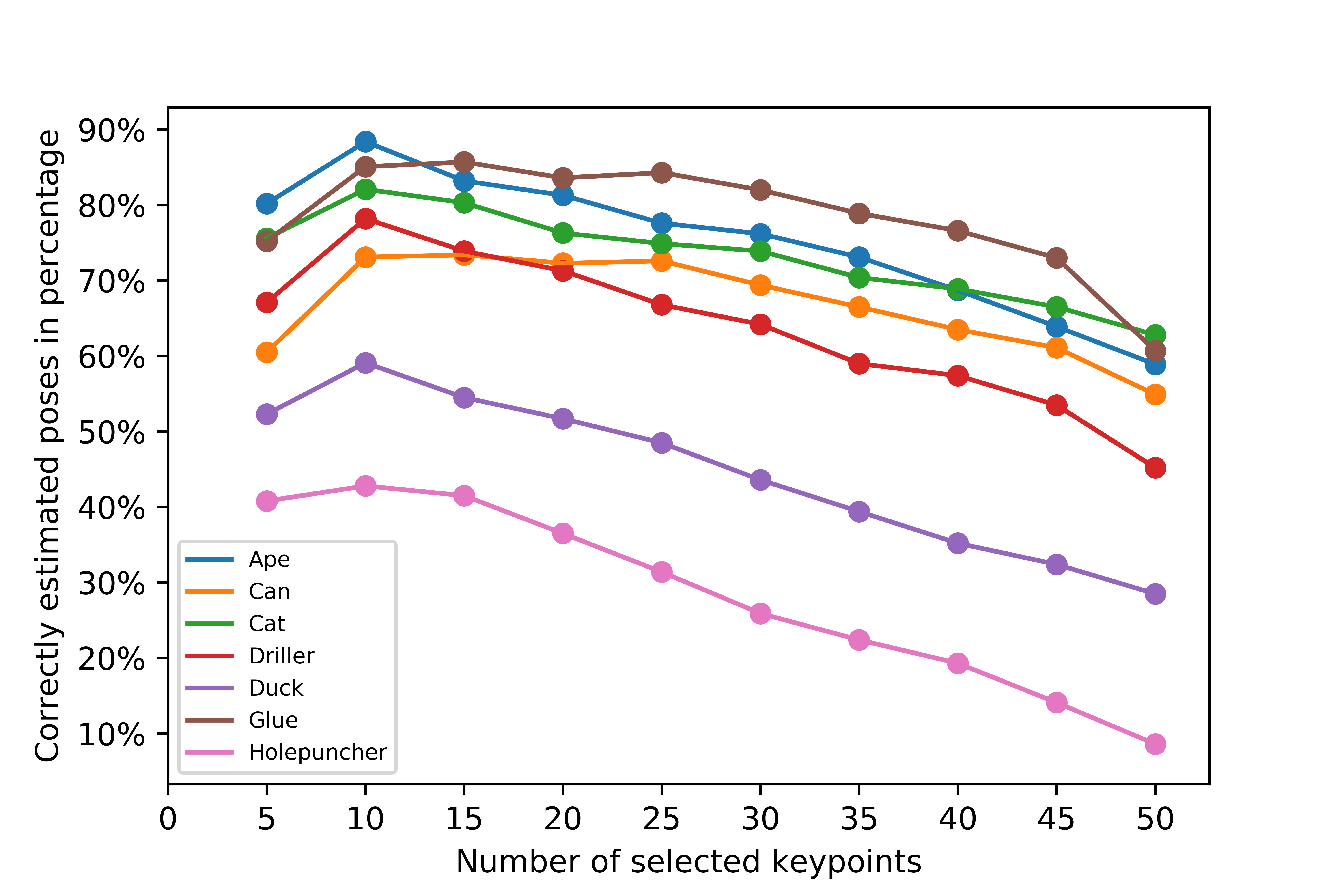}
\end{center}
   \caption{The relation between 2D reprojection accuracy of predicted poses and number of selected most confident keypoints which are used in PnP algorithm.}
\label{fig:number_of_selected_keypoints}
\end{figure}

In theory, the PnP algorithm needs at least three pairs of keypoints to recover an object's 6D pose. We can designate much more than three keypoints and thus being free to select the most confident keypoints to overcome occlusion. As we mentioned in Section \ref{sec:occlusionresult}, among $50$ keypoints, we select top-$10$ confident estimated keypoints and sacrifice others. We visualize this procedure in Figure \ref{fig:filter10}. We also conduct a quantitative experiment to show the relation between the pose estimation accuracy and the number of selected most confident keypoints. The results are in Figure \ref{fig:number_of_selected_keypoints}. We can see that selecting the top-10 confident keypoints can lead to prominently more accurate poses in comparison with preserving all 50 predicted keypoints. Generally selecting fewer most-confident keypoints can lead to better 6D pose estimation because the selected keypoints are relatively more accurate. However, few selected keypoints would do harm to the effectiveness of the PnP algorithm.

\subsection{Surface Keypoints Designation}
\label{sec:choosing keypoints}
We designate the 3D SIFT surface points as keypoints instead of using 8 corners and the center of 3D bounding boxes as \cite{Rad2017BB8AS,tekin18} do. And we conduct a comparative experiment to validate that SIFT keypoints are better than corners points. We only change $M_{k3D}$, the set of keypoints to conduct all the comparative experiments based on our pipeline. (Note that in our pipeline design, points in $M_{k3D}$ aren't necessary to be on the model surface.) For the sake of fairness, we restrict the number of keypoints to 9, and we keep all the training settings the same. We train each of them for enough epochs to ensure they both converge well.

\begin{figure}[t]
\begin{center}
   \includegraphics[width=\linewidth]{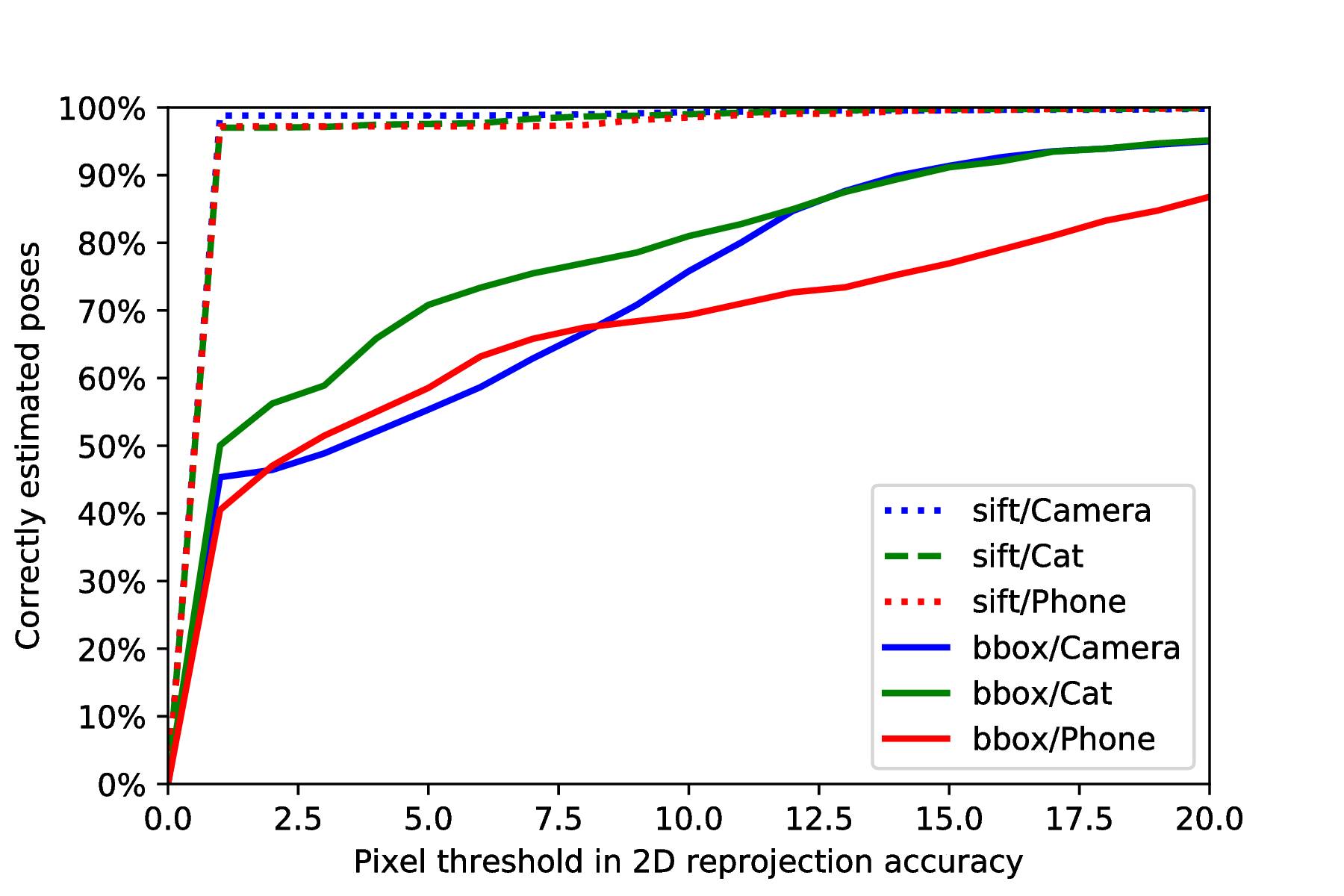}
\end{center}
   \caption{Comparison between two ways of keypoint designation: SIFT points and the corners plus center of bounding boxes. We test on three objects on LineMod dataset in terms of 2D projection accuracy.}
\label{fig:diff_ways}
\end{figure}

The overall results are in Figure \ref{fig:diff_ways}. We can see that choosing surface points as keypoints works much better than corners and the center of the 3D bounding box. This is unsurprising because the surface keypoints are more related to the features of the target results than the bounding box vertices in the air and thus the network can estimate the keypoints more accurately. This comparison can partially explain why our pipeline achieves better results on LineMod than \cite{Rad2017BB8AS,tekin18}.

\section{Conclusion}
We have proposed a novel and natural method to estimate 6D pose mainly by localizing the designated surface keypoints. Several experiments are conducted to validate the accuracy, speed and robustness of our method. Concerning accuracy, our approach surpasses the state-of-the-art not-using-refinement method by a large margin and is competitive with the state-of-art using-refinement method. Meanwhile, our method is much faster than using-refinement methods. Additionally, our approach can handle symmetry naturally and is robust to occlusion. We are looking forward to future work about better ways of keypoint designation to recover 6D pose more accurately.



{\small
\bibliographystyle{ieee}
\bibliography{egbib}
}
\end{document}